\documentclass[10pt]{IEEEtran}

\usepackage{pstricks, pst-plot}	% pstricks
\usepackage{graphicx} % enables import of various formats
\usepackage{wrapfig}  % enables floating wrapped figures
\usepackage[figuresright]{rotating}
\usepackage{multicol}

\usepackage{amsmath}
\usepackage{amssymb}
\usepackage[cal = txupr]{mathalpha}
\usepackage[english]{babel}
\usepackage{blindtext}
\usepackage[ruled]{algorithm2e}
\usepackage{caption}
\usepackage{changepage}
\usepackage{capt-of}
\usepackage{hyperref}
\usepackage{breqn}
\hypersetup{
    colorlinks=black,
    linkcolor=black,
    filecolor=black,      
    urlcolor=black,
    pdftitle={Overleaf Example},
    pdfpagemode=FullScreen,
    }

\usepackage{amsthm}
\newtheorem{theorem}{Theorem}

\newtheorem{proposition}[theorem]{Proposition}

\title{Distributional Constrained Reinforcement Learning for Supply Chain Optimization}
\author{
    \IEEEauthorblockN{Jaime Sabal Bermúdez$^{a}$, Antonio del Rio Chanona$^{b}$, Calvin Tsay$^{a,b,*}$\thanks{$^*$c.tsay@imperial.ac.uk}}\\
    \IEEEauthorblockA{$^{a}$Department of Computing, Imperial College London
    }\\
    \IEEEauthorblockA{$^{b}$Sargent Centre for Process Systems Engineering, Imperial College London}
}

\begin{document}

\maketitle             % make title page with abstract and keywords

% ================================================================================

\begin{abstract}
This work studies reinforcement learning (RL) in the context of multi-period supply chains subject to constraints, e.g., on production and inventory. 
We introduce Distributional Constrained Policy Optimization (DCPO), a novel approach for reliable constraint satisfaction in RL. 
Our approach is based on Constrained Policy Optimization (CPO), which is subject to approximation errors that in practice lead it to converge to infeasible policies. 
We address this issue by incorporating aspects of distributional RL into DCPO. 
Specifically, we represent the return and cost value functions using neural networks that output discrete distributions, and we reshape costs based on the associated confidence.  
Using a supply chain case study, we show that DCPO improves the rate at which the RL policy converges and ensures reliable constraint satisfaction by the end of training. 
The proposed method also improves predictability, greatly reducing the variance of returns between runs, respectively; this result is significant in the context of policy gradient methods, which intrinsically introduce significant variance during training.
\end{abstract}
% ==================================== body  =====================================

\section{Introduction}
Recent years have highlighted the importance of efficient supply chains in the development and functioning of modern society. 
Here, the field of inventory management deals with optimization of the ordering, storing, and selling of product inventory. 
There are many aspects of this problem that make it challenging, and a lot of benefits in doing so that make it worth investigating.
Challenges include the presence of outdated ``legacy'' processes or the inefficient use of available physical capital. A variety of operations research methods have been used in tackling this family of problems, including dynamic programming, linear programming, and game theory. Given complex environments, however, it is clear that modern inventory management also lends itself towards optimization through reinforcement learning (RL).

In the past few decades, machine learning has gained increasing popularity for its capacity to imitate the way humans learn, mimicking aspects of cognitive reasoning. 
RL provides a framework for artificial agents to learn by interacting with an external environment and iteratively updating the way in which they act (i.e., their \textit{policy}). Specifically, each interaction moves the agent to a new state and produces some quantifiable \textit{reward}; the agent seeks to maximize cumulative reward (i.e., \textit{return}) from a series of interactions. RL has proven to be very effective in areas such as gaming, demonstrating superhuman performance in incredibly complex games, or robotics, control~\cite{nian2020review, shin2019reinforcement}, and scheduling~\cite{ikonen2020reinforcement, shyalika2020reinforcement} where problems can be too complex for an analytical solution. 

Real-world environments such as supply chains, may impose physical or safety limitations on the actions an agent can take.  
This work introduces a framework for RL in these circumstances, combining advantages of constrained and distributional RL. 
\textit{Constrained RL} algorithms extend the objective of an RL agent to include minimising the expected costs below some threshold defined by the constraints in an environment, in addition to maximising expected returns~\cite{garcia2015comprehensive}. 
In a related vein, \textit{distributional RL} attempts to imitate the way in which organisms consider risk in decision-making, by estimating the distribution of values for all the states in the environment~\cite{distributional-perspective}. The intrinsic stochasticity of many environments motivates using probability distributions to evaluate the possible future rewards associated with an action \cite{petsagkourakis2022chance}. By estimating these distributions, distributional RL simplifies the parameterization of risk-adversity in RL, giving benefits in problems where constraints should be satisfied under uncertainty.

\section{Reinforcement Learning (RL) Background}

RL often models agent-environment interaction as a Markov Decision Process (MDP). An MDP is defined by the tuple $\left(\mathcal{S}, \mathcal{A}, \mathcal{R}, \mathcal{P}, \gamma\right)$, where $\mathcal{S}$ is the set of states in the environment, $\mathcal{A}$ is the set of possible actions, $\mathcal{R}: \mathcal{S} \times \mathcal{A} \times \mathcal{S} \rightarrow \mathbb{R}$ is a reward function, $\mathcal{P}: \mathcal{S} \times \mathcal{A} \times \mathcal{S} \rightarrow \left[0,1\right]$ is the transition probability function ($P\left(s' \mid s,a\right)$ is the probability of transitioning to state $s'$ by taking action $a$ from state $s$), and $\gamma$ is a discount rate for future rewards. 
A policy $\pi: \mathcal{S} \rightarrow \mathcal{P}\left(A\right)$ maps states to probability distributions over the possible actions ($\pi\left(a \! \mid \! s\right)$ is the probability of taking action $a$ from state $s$). We denote the set of all possible stationary policies as $\Pi$. 
RL seeks a policy $\pi$ that maximizes a performance measure $J_{R}\left(\pi\right)$, often taken as $J_{R}\left(\pi\right) =\mathbb{E}_{\tau \sim \pi} [\sum_t^\infty \gamma^t R(s_t, a_t, s_{t+1})]$, where $\tau = \left(s_0, a_0, s_1, ... \right)$ is a trajectory sampled using $\pi$. Denoting $\hat{R}\left(\tau\right)$ as the discounted return of a trajectory, the on-policy state value function is denoted $V^\pi (s) = \mathbb{E}_{\tau \sim \pi}\left[\hat{R}\left(\tau\right) \mid s_0 = s\right]$, with optimal policy $\pi^* = \mathrm{argmax}_{\pi} V^\pi \left(s\right)$. Analogously, the state-action value function is defined as $Q^\pi \left(s,a\right)=\mathbb{E}_{\tau \sim \pi}\left[\hat{R}\left(\tau\right) \mid s_0 = s, a_0 = a\right]$ and the advantage function of an action is $A^\pi \left(s,a\right) = Q^\pi \left(s,a\right) - V^\pi \left(s\right)$.

Traditional reinforcement learning algorithms suffer from the \textit{curse of dimensionality} \cite{bellman-dp}. This phenomenon describes the challenge of obtaining reliable results when dealing with a high-dimensional state space, arising from the fact that the amount of data needed to achieve a result increases exponentially with dimensionality. 
Given this challenge, \textit{deep RL} employs a nonlinear function approximator, i.e., a deep neural network (NN) with parameters $\theta$, to estimate the optimal state-action value function as $Q(s,a; \theta) \approx Q^*(s,a)$ for very complex and large state and action spaces. Alternatively, we can also learn a parameterised function $\pi_\theta(a  \mid s)$ for the policy directly which allows for the optimization of problems with continuous action spaces. In policy gradient methods, we learn a parameterised function for the policy directly, written as $\pi_\theta(a \! \mid \! s)$, in terms of $\theta$. To do this, a neural network is trained to maximise the objective $J\left(\theta\right)$, which measures the expected return from following $\pi_\theta$. The policy gradient theorem \cite{policy-gradients} provides an expression for the derivative of the objective with respect to the policy parameters $\theta$: 

\begin{equation}
    \nabla_\theta  J(\theta) = \mathbb{E}_\pi \left[Q^\pi(s,a) \nabla_\theta \ln \pi_\theta(a \! \mid \! s) \right]
\end{equation}

\noindent which results in an update rule that is unbiased but has a large variance in the gradient estimates. The latter is commonly alleviated through the use of the advantage in place of $Q^\pi(s,a)$; estimated through Generalised Advantage Estimation (GAE) \cite{GAE} by using $V^\pi(s)$ as a baseline.

%DEEP RL
\vspace{0.1in}
\textbf{Trust Region Policy Optimization.}
A more recent strategy to deal with noisy gradient estimates is Trust Region Policy optimization (TRPO)~\cite{trpo}, which limits the Kullback-Leibler (KL) divergence between the current policy $\pi$ and the previous policy $\pi_k$ to a threshold $\delta$. Training the policy parameters $\theta$ thus involves: 
\begin{equation}
\begin{aligned}
&\underset{\theta}{\text{max}} &\mathbb{E}_{s \sim \rho^{\pi_k}, a \sim \pi_k} \left[\frac{\pi (a \! \mid \! s)}{\pi_k (a \! \mid \! s)} A^{\pi_k} \left(s,a\right)\right] \\
&\text{s.t.} &\mathbb{E}_{s \sim \rho^{\pi_k}} \left[ \overline{D}_{\mathrm{KL}} \left(\pi_k \left(\cdot \! \mid \! s\right) \mid \mid \pi \left(\cdot \! \mid \! s\right) \right) \right] \leq \delta
\label{eq:TRPO}
\end{aligned}
\end{equation}
where $\rho^{\pi_k}$ is the state visitation probability density under $\pi_k$, and $\overline{D}_{\mathrm{KL}}$ denotes the average KL-divergence. %Note that here we express $\pi_\theta$ and $\pi_{k,\theta}$ as $\pi$ and $\pi_k$, respectively, for simplicity. 
The problem \eqref{eq:TRPO} is typically reformulated using Monte-Carlo estimates of $Q^{\pi_k}(s,a)$, a linearization of the objective, and a second-order approximation of the trust-region constraint.
The reformulated problem is typically then solved using conjugate gradients and a backtracking line search. 

%CMDPs
\vspace{0.1in}
\textbf{Constrained Reinforcement Learning.} 
A natural extension considers RL in settings where the environment is constrained. 
Constrained RL algorithms often assume a constrained MDP (CMDP), which models costs $C: \mathcal{S} \times \mathcal{A} \times \mathcal{S} \rightarrow \mathbb{R}$ analogously to rewards, as well as a threshold $d$. The objective is to maximise $J_{R}\left(\pi\right)$ such that a cost constraint $J_{C}\left(\pi\right) = \mathbb{E}_{\tau \sim \pi} [\sum_t^\infty \gamma^t C(s_t, a_t, s_{t+1})] \leq d$ is satisfied.
This can be expressed in terms of the optimal policy $\pi^* = \mathrm{argmax}_{\pi \in \Pi_C} J_R \left(\pi\right)$, where $\Pi_C$ $\dot{=}$ $\{\pi \in \Pi : J_C\left(\pi\right) \le d\}$ is the set of cost-feasible policies. 

Incorporation of this constrained setting is non-trivial for DQN-based methods, where a common strategy is to penalize the reward function with constraint violation(s)~\cite{yoo2022dynamic}. 
On the other hand, incorporating environment constraints in a policy update strategy is more straightforward, as constraints can be incorporated into an existing optimization problem, e.g., \eqref{eq:TRPO}. 

\textit{Constrained Policy Optimization} (CPO)~\cite{cpo} restricts policy updates similarly to TRPO, but to satisfy both the trust-region constraint from \eqref{eq:TRPO} and environment constraints. 
The proposed optimization problem can be solved efficiently using duality, where Lagrange multipliers for both the objective and environment constraints are estimated at each step. 
Another way of enforcing constraints is to augment the state-space and reshape the objective of the problem. \textit{Safety Augmented (SAUTE) MDPs}~\cite{saute} allow for a reformulation of the problem in terms of  minimisation of the reshaped objective, whereby safety constraints are incorporated through a safety budget. 
Once this budget is exhausted the objective takes on an infinite value, leading to constraint satisfaction with probability equal to one by the end of training. 
This approach only modifies the environment and is compatible with most off-the-shelf RL algorithms, but the state space increases with the number of environment constraints, leading to scalability issues and thus a loss in sampling efficiency if the number of constraints is large.

\section{Approach: Distributional Constrained Policy Optimization}
In this work we show that applying distributional RL to the CPO setting can strike a balance between reliable constraint satisfaction and consistent sampling efficiency, independent of the number of constraints.
Distributional RL~\cite{distributional-perspective} extends traditional RL to consider the probability distribution of returns over all state-action pairs. This approach provides a range of auxiliary statistics, such as the variance in returns, that together provide a more natural framework for inductive bias. 
In a similar manner, we propose to approximate the cost distribution over state-action pairs to manage risk aversion during training. 
Pseudo-code for this distributional version of CPO, which we call DCPO, is given in Algorithm \ref{alg: dcpo_algo}.  

\vspace{0.1in}
\textbf{Problem Formulation. }
Consider the CPO policy update with a single environment constraint: 
\begin{equation} \label{eq:cpo_objective}
\begin{aligned}
&\underset{\pi \in \Pi_\theta}{\text{max}} &&\mathbb{E}_{s \sim \rho^{\pi_k}, a \sim \pi_k} \left[A^{\pi_k} \left(s,a\right)\right] \\
&\text{s.t} &&J_{C}(\pi_k) + \frac{1}{1 - \gamma} \mathbb{E}_{s \sim \rho^{\pi_k}, a \sim \pi_k} \left[A_{C}^{\pi_k}(s,a)\right] \le d \\
& &&\overline{D}_{\mathrm{KL}} \left(\pi_k \mid \mid \pi\right) \leq \delta \\
\end{aligned}
\end{equation}
\noindent where $A_{C}^{\pi_k}$ denotes the estimated cost advantage under $\pi_k$ for the cost function $C$. 
This update rule is computationally expensive for high-dimensional NN parameter spaces. For small $\delta$, we can linearize the objective and safety constraints around $\pi_k$. Denoting the gradients of the objective and constraint as $g$ and $b$, respectively, the Hessian of the KL-divergence as $H$ (i.e., Fisher Information Matrix), and defining $c = J_C\left(\pi_k\right) - d$, \eqref{eq:cpo_objective} can be written in terms of the step direction $x = \theta - \theta_k$: 

\begin{equation} \label{eq:cpo_approximation}
\begin{aligned}
& p^* = && \underset{x}{\text{min \hspace{0.05cm}}} g^T x \\
& \text{s.t} && c + b^T x \leq 0,\ x^T H x \leq \delta
\end{aligned}
\end{equation}

\noindent with $g,b,x \in \mathbb{R}^n$, $c,d \in \mathbb{R}$, $\delta > 0$, and $H \succ 0$. 
We note that $g$ and $b$ can be computed by back-propagation, and that $H$ is always positive semi-definite, making \eqref{eq:cpo_approximation} a convex optimization problem that can be solved be solved through KKT stationary conditions when feasible. 
Given a single cost function, the optimal point $x^*$ (assuming one exists) is 
\begin{equation}
x^* = -\frac{1}{\lambda^*} H^{-1} \left(g - \nu^* b\right)
\end{equation}

The dual variables $\lambda^*$ and $\nu^*$ (Lagrange multipliers) are: 
\begin{equation} \label{eq:opt_duals}
\begin{aligned}
\nu^* & = \left(\frac{\lambda^* c - r}{s}\right)_+, \\
\lambda^* & =  \underset{\lambda \geq 0}{\text{argmax}} 
\begin{cases}
    f_a\left(\lambda\right) & \text{if} \hspace{0.2cm} \lambda c - r > 0 \\
    f_b \left(\lambda\right) & \text{otherwise},
\end{cases} \\
\end{aligned}
\end{equation}
where $q = g^T H^{-1} g$, $r = g^T H^{-1} b$, $s = b^T H^{-1} b$, and  $f_a\left(\lambda\right) \doteq \frac{1}{2 \lambda} \left(\frac{r^2}{s} - q\right) + \frac{\lambda}{2} \left(\frac{c^2}{s} - \delta\right) - \frac{rc}{s}$ and $f_b \left(\lambda\right) \doteq -\frac{1}{2 \lambda} \left(\frac{q}{\lambda} + \lambda \delta\right)$.
In practice $H^{-1}$ is expensive to compute so we approximately solve for $H^{-1} g$ and $H^{-1} b$ using conjugate gradients. Since $\lambda \geq 0$, we must restrict the optimal $\lambda$ from the two cases above through a projection to the sets $\Lambda_a \doteq \{\lambda \mid \lambda c - r > 0, \lambda \geq 0\}$ and $\Lambda_b \doteq \{\lambda \mid \lambda c - r \leq 0, \lambda \geq 0\}$, respectively, such that 
\begin{equation}
\lambda^* \in \Biggl\{\text{Proj}\left(\sqrt{\frac{q - r^2 / s}{\delta - c^2 s}}, \Lambda_a\right), \text{Proj}\left(\sqrt{\frac{q}{\delta}}, \Lambda_b\right)\Biggr\}
\end{equation}

\begin{equation}
    \lambda^* = \begin{cases}
        \lambda_a^* & \text{if} \hspace{0.2cm} f_a\left(\lambda_a^*\right) \geq f_b\left(\lambda_b^*\right) \\

        \lambda_b^* & \text{otherwise}.
    \end{cases}
\end{equation}

It is worth mentioning that $H$ is computed using the quadratic constraint such that any step taken satisfies the KL-divergence constraint. In the infeasible case when $c^2 s - \delta > 0$ and $c > 0$, we apply a recovery method towards constraint satisfaction, $x^* = - \sqrt{2 \delta/(b^T H^{-1} b)} H^{-1} b$.

\vspace{0.1in}
\textbf{Distributional Value Function and Safety Baseline.} 
Accurate estimates of the return and safety advantages $A^{\pi_k}$ and $A_{C}^{\pi_k}$ in \eqref{eq:cpo_objective} are essential to reduce the variance associated with the policy-gradient update step. 
These advantage functions can be estimated, e.g., through Generalized Advantage Estimation (GAE) \cite{GAE}, and approximated with NNs.  
To model uncertainty, we choose to represent the return and cost value functions (used in estimating $A^{\pi_k}$ and $A_{C}^{\pi_k}$) using NNs whose outputs are distributions parameterized by $N$ equally spaced quantiles in a range $\left[V_{\text{min}}, V_{\text{min}}\right]$ \cite{distributional-perspective}. We employ a surrogate loss for these NNs comprising the average negative log-likelihood of obtaining the costs or returns sampled under policy $\pi_k$. To simplify computation of the log-likelihood, we approximate the quantile distribution as Gaussian. For a value function with parameters $\phi$ the loss is thus $\mathcal{L}\left(\phi\right) = - \log\left(p\left(J_R\left(\pi_k\right) \! \mid \! \phi \right)\right)$, which is trainable using stochastic gradient descent. 

\vspace{0.1in}
\textbf{Cost Reshaping. }
To reliably satisfy constraints, we reshape the cost $J_C\left(\pi_k\right)$ and safety threshold $d$ based on the agents' confidence that constraints are satisfied given a cost distribution with parameters $\phi_c$. 
Specifically, we introduce a reshaping parameter $\rho$, whose value is set to $\beta \mathbb{P}  \left(J_{C}^{\pi_k} > d; \phi_c\right)$ when $J_{C}^{\pi_k} > d$ (i.e. the constraints are violated), and to $- \beta \mathbb{P}  \left(J_{C}^{\pi_k} < d; \phi_c\right)$ when the constraints are satisfied, with $\beta \geq 0$. 
Note that we evaluate $J_C$ under the previous policy $\pi_k$. 
In other words, $\rho$ takes the value of $\beta$ when the agent is fully confident that constraints are violated under the previous policy $\pi_k$, and $-\beta$ when it is fully confident the constraints are satisfied. We then reshape $J_C^{\pi_k}$ and $d$ as $\tilde{J}_{C}^{\pi_k} = J_C^{\pi_k} \left( 1 + \textrm{clip} \left( \rho, -\epsilon, \infty \right) \right)$ and $\tilde{d} =  d  \left( 1 + \textrm{clip} \left( \rho, -\epsilon, \infty \right) \right)$.
The new parameter $\epsilon$ is a safety margin that controls risk aversion. 
This reshaping only affects policy updates when constraints are satisfied, effectively avoiding large steps in directions that possibly increase costs. Intuitively, maximising rewards in a constrained environment usually leads to the agent to the edges of the feasible region.  
We thus mitigate risk by adjusting the feasibility of the trust region depending on the confidence of constraint satisfaction, while maintaining the same optimisation problem as CPO (see Proposition \ref{theorem}).

\begin{proposition} \label{theorem}
    For any functions $f : S \rightarrow \mathbb{R}^+$, $p : S \rightarrow \mathbb{R} \in \left[-1,1\right]$, and with $d,\beta \in  \mathbb{R}^+$, $\epsilon \in \left(0,1\right)$, define
    \begin{align}
        k(s) & = \begin{cases}
            -\epsilon & \text{if} \hspace{0.2cm} \beta p(s) < -\epsilon, \\
            \beta p(s)  & \text{otherwise}, \\
        \end{cases}
    \end{align}
    \vspace{-0.1cm}
    If we multiply $f(s)$ and $d$ by the scalar $1 + k(s)$ as,
    \begin{align}
        \tilde{f}(s) & = f(s)\left(1 + k(s)\right) \\ 
        \vspace{0.2in}
        \tilde{d}(s) & = d\left(1 + k(s)\right)
    \end{align}
    \vspace{-0.1cm}
    Then, 
    \begin{equation} \label{conclusion}
        f(s) - d > 0 \implies \tilde{f}(s) - \tilde{d}(s) > 0 
    \end{equation}
\end{proposition}

\vspace{-0.1in}
\begin{proof}  
$1 + k(s) > 0$ given $\epsilon \in \left(0,1\right)$ and $\beta \in \mathbb{R}^+$. Therefore, this immediately holds. 
\end{proof}

\setlength{\textfloatsep}{4pt}

\begin{algorithm*}[!t]
\caption{Distributional Constrained Policy Optimization (DCPO)}
\textbf{Input:} {Initial policy $\pi_0 \in \Pi_\theta$, safety margin $\epsilon$, reshaping coefficient $\beta$} \\
 \For{$k=0,1,2,...$}{
  Sample a set of trajectories $\mathcal{D} = {\tau} \sim \pi_k = \pi\left(\theta_k\right)$\\
  %Reshape ${J}_{C}\left(\pi_k\right)$ and target ${d}$ \\
  Form sample estimates $\hat{g}$, $\hat{b}$, $\hat{H}$, $\hat{c}$ using $\mathcal{D}$, and reshaped $\tilde{J}_{C}\left(\pi_k\right), \tilde{d}$\\
  \eIf{approximate CPO \eqref{eq:cpo_approximation} is feasible}{
   Solve dual problem with reshaped constraint $\tilde{c} = \tilde{J}_C\left(\pi_k\right) - \tilde{d}$\\
   Compute new policy $\theta^\ast$ using $x^* = -\frac{1}{\lambda^*} H^{-1} \left(g - \nu^* b\right)$
   }{
   Compute recovery policy $\theta^\ast$ using $x^* = - \sqrt{2 \delta/(b^T H^{-1} b)} H^{-1} b$
  } 
  Obtain $\theta_{k+1}$ through backtracking linesearch to enforce constraints in \eqref{eq:cpo_objective} over $\mathcal{D}$
 }
 \label{alg: dcpo_algo}
\end{algorithm*}

\section{Results and Discussion}

\subsection{Supply Chain Case Study}
We consider the inventory management problem for a multi-echelon, multi-period supply chain~\cite{inventory-management}.
Supply chains consist of a a series of interconnected nodes at the different stages of production that aim to turn a raw product into one that is ready for consumption at the final stage of the pipeline. 
The operational goal is to maximise profits while satisfying demand and inventory constraints. 
We consider the \textit{InvManagement-v0} problem in the \verb|or-gym| library~\cite{HubbsOR-Gym}. 
The sequence of events in a given period can be summarised as:

\begin{enumerate}
    \item Main network nodes place replenishment orders to their suppliers (located at the respective preceding stage). Orders are filled according to the suppliers' available production capacity and available inventory. 

    \item Main network nodes receive inventory replenishment shipments that have made it down the pipeline after the associated lead times have transpired, which take into account both production and transportation times. 

    \item Single-product customer demand is filled according to the available inventory at the retail node. 

    \item Unfulfilled sales are backlogged at a cost and subsequently take priority in the following period, where they are sold at a lower profit. 

    \item Surplus inventory is held at each stage at a holding cost. 

    \item Remaining inventory after the final period is lost. 
\end{enumerate}

Mathematically, the dynamics of the environment can be expressed for each period $t$ and $\forall m \in \mathcal{M}, t \in \mathcal{T}$: 
\begin{subequations}
\begin{align}
I_{t+1}^m & = I_t^m + R_{t-L_m}^m - S_t^m  \\
T_{t+1}^m & = T_t^m - R_{t-L_m}^m + R_t^m \\
S_t^m & =  \begin{cases}
            R_t^{m-1} & \text{if} \hspace{.2cm} m > 0, \\
            \min\left(I_t^0 + R_{t-L_m}^0, D_t + B_{t-1}^0\right) & \text{if} \hspace{.2cm} m = 0
        \end{cases} \\
U_t^m & = R_t^{m-1} - S_t^m \\
P_t^m & = \left(\alpha\right)^n \left(p^m S_t^m - r^m R_t^m - k^m U_t^m - h^m I_{t+1}^m\right)
\end{align}
\end{subequations}
Eqs. 13a--b are mass balances for the on-hand ($I$) and pipeline ($T$) inventories for each stage $m$, where $R$ and $L$ denote replenishment orders and lead times, respectively. 
Eq. $\text{13c}$ gives the sales ($S$) for the retailer ($m = 0$) and the rest of the pipeline ($m > 0$), given demand $D$ and backlog $B$. 
Eq. $\text{13d}$ describes that the unfulfilled re-order quantities ($U$), and Eq. $\text{13e}$ calculates the profit $P$ as the sales revenue minus procurement costs, unfulfilled demand penalties, and excess inventory penalties, all subject to a discount factor $\alpha = 0.97$. The unit sales price $p$, and penalties $k, h$ are all known. 

There are production capacity ($c^{m+1}$) and inventory ($I_t^{m+1}$) constraints on the re-order quantity $R_t^m$. To enforce these, we neglect the explicit enforcement done in \cite{HubbsOR-Gym} and alternatively define a cost function that sums the number of constraint violations $C\left(s, a\right) = \sum_{m \in \mathcal{M},t \in \mathcal{T}} C\left(I_t^{m+1}, R_t^m\right)$, where $C\left(I_t^{m+1}, R_t^m\right)$ is the number of constraints violated at node $m$ at time $t$. 
We use this indicator cost function rather than a continuous one in order to limit the number of production capacity and inventory, regardless of their degree. 
The objective of our agent is then to maximize profits with a limit on the number of production capacity and inventory constraint violations. 

\subsection{DCPO Results}
\begin{figure}[h] 
    \centering
    \includegraphics[width=0.45\textwidth]{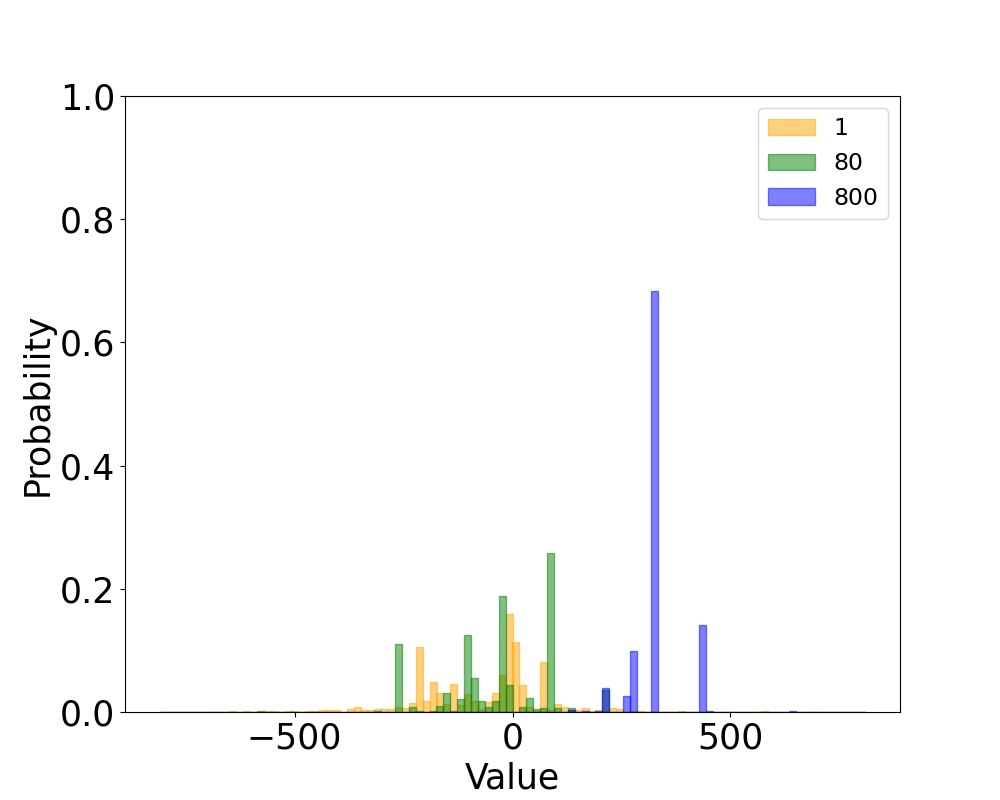}
\caption{Learned quantiles ($N=102$) for the return distribution of the initial state $s_0$ at epochs 1, 80, and 800 in training.}
\label{fig:quantiles-dist}
\end{figure}

All experiments were ran with ten random seeds and hyperparameters were hand-tuned through trial and error for each algorithm to obtain reasonable results in a sensible amount of time. Moreover, all of the algorithms use separate feed-forward NN's for the policy, the return value distribution, and the cost value distribution of size (64, 64) with tanh activations. Running DCPO \footnote{\href{http://github.com/jaimesabalimperial/jaisalab}{{http://github.com/jaimesabalimperial/jaisalab}}} with $N = 102$ quantiles shows the convergence of the learned return distributions during training (Figure \ref{fig:quantiles-dist}). %Indeed, the progression of the quantiles aligns with the theory regarding their uncertainty. 
Intuitively, at earlier periods the agent has seen fewer transitions, and is therefore uncertain about the cost/value of a state (epistemic uncertainty) \cite{risk-uncertainty}. 
This is evident in the relatively flat learned distributions at epochs 1 and 80. 
As more samples are observed, the distribution converges around a few individual quantiles, and remaining uncertainty in the value of a state likely stems from the stochasticity of the MDP (aleatoric uncertainty). 
We observe a similar trend for the learned cost distributions during training. 

\begin{figure}[hbtp]
  \begin{center}
    \includegraphics[width=\columnwidth]{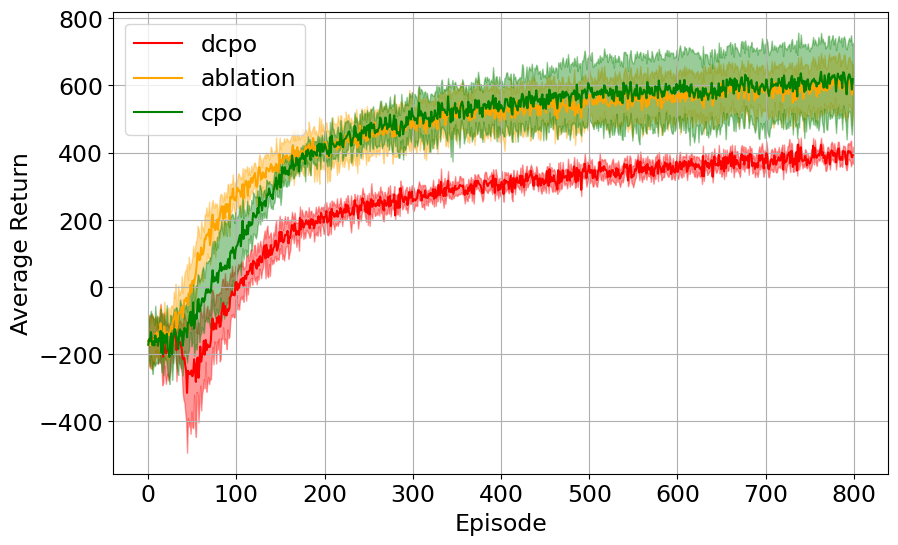}
    \hspace*{0cm}
    \includegraphics[width=\columnwidth]{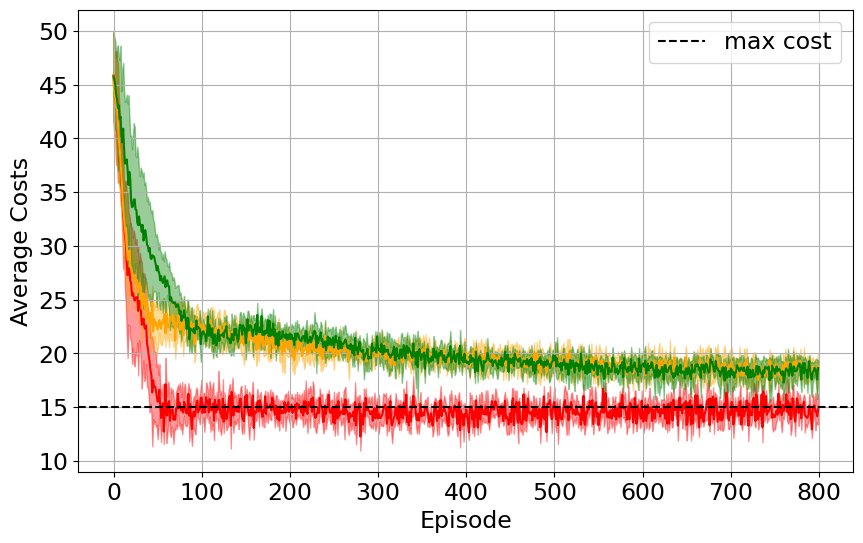}
  \end{center}
  \caption{Returns (top) and costs (bottom) incurred by CPO, DCPO, and DCPO without cost reshaping (`ablation') over 5 runs. DCPO ($\beta = 100.0$, $\epsilon=0.15$) converges to the cost limit. }
  \label{fig:ablation}
\end{figure}

\vspace{0.1in}
\textbf{Importance of Cost Reshaping. }
We first consider an ablation experiment on the reshaping of the costs to examine the isolated effect of applying distributional RL to CPO. 
Figure \ref{fig:ablation} shows the costs and returns incurred during training for CPO, DCPO, and `ablation' experiments (DCPO without cost reshaping). We observe that using parametric return and cost value distributions improves the initial stages of training. Specifically, the `ablation' agent learns more quickly to both minimize costs and maximize returns, while converging to similar values as CPO. 
Moreover, this also reduces the standard deviation of the returns across replications, with a $40.0\%$ compared to CPO at the end of training. This effect is even stronger when the constraint is reshaped, with DCPO having a $76.6\%$ decrease compared to CPO. 

\begin{figure}[h]
  \begin{center}
    \includegraphics[width=\columnwidth]{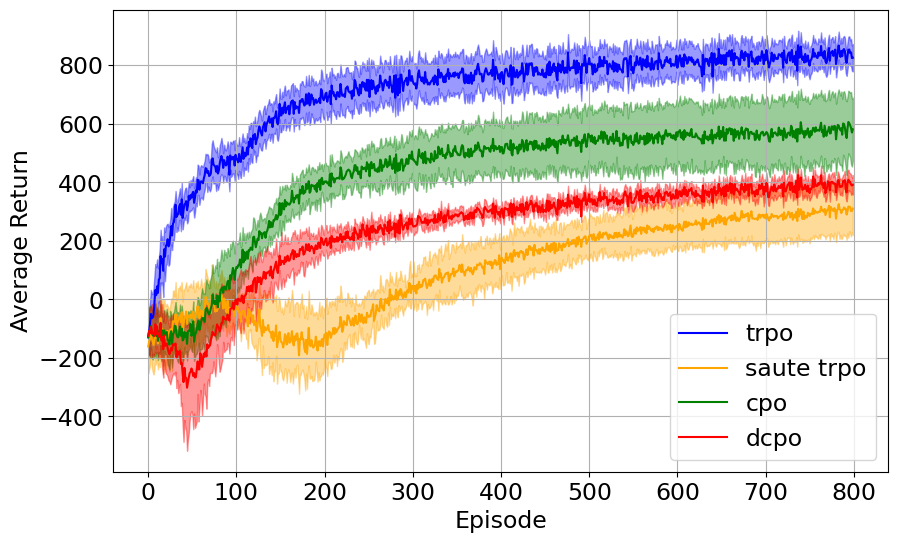}
    \hspace*{-0cm}
    \includegraphics[width=\columnwidth]{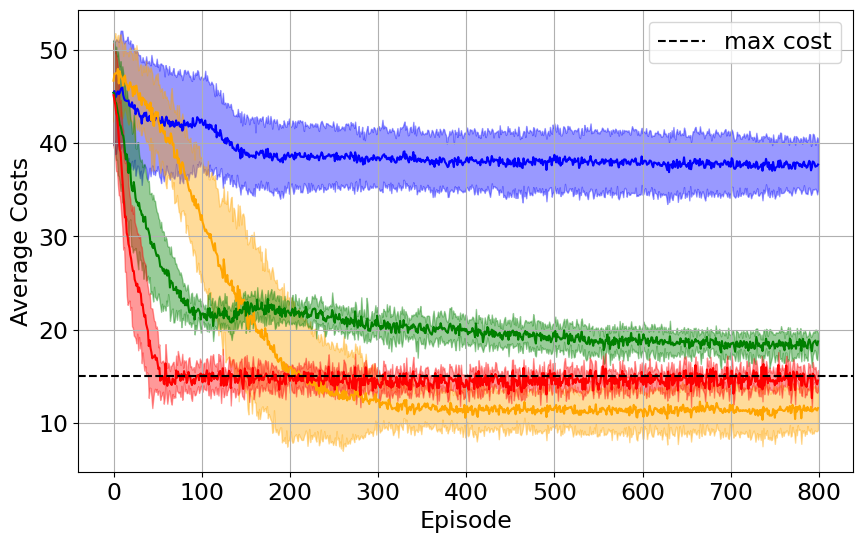}
  \end{center}
  \caption{Returns (top) and costs (bottom) for TRPO, Saute TRPO, DCPO, and CPO over 10 runs.}
  \label{fig:final_training}
\end{figure}

\vspace{0.1in}
\textbf{Comparison to TRPO and Saute. }
Figure \ref{fig:final_training} shows the costs and returns incurred during training for TRPO, Saute TRPO, CPO, and DCPO. 
TRPO is unconstrained and produces infeasible policies, more than doubling the permitted number of violations, but it attains the largest returns. 
On the other hand, Saute TRPO incurs costs well below the limit by the end of training, with a clear trade-off in the low returns. 
DCPO seems to balance the intrinsic trade-off between returns and costs, with mean costs below the defined limit, but with higher returns than Saute TRPO. 
Moreover, DCPO showed a reduction of $70.8\%$ in the standard deviation of the returns relative to TRPO by the end of training, which may be attributed to the risk aversion intrinsic to the reshaping of the constraint. Specifically, by the end of training, constraints are consistently satisfied, and steps in a constraint-violating direction will produce a subsequent step that aggressively draws it back to a feasible part of the trust region. 
This largely limits the policy updates to a region around the exact cost threshold, as the given environment exhibits a clear trade-off between returns and costs. 
Finally, CPO converged to intermediate values for both costs and returns, potentially owing to approximation and sampling errors in the objective/cost gradients. 

\begin{figure}
\centering
  \includegraphics[width=\columnwidth]{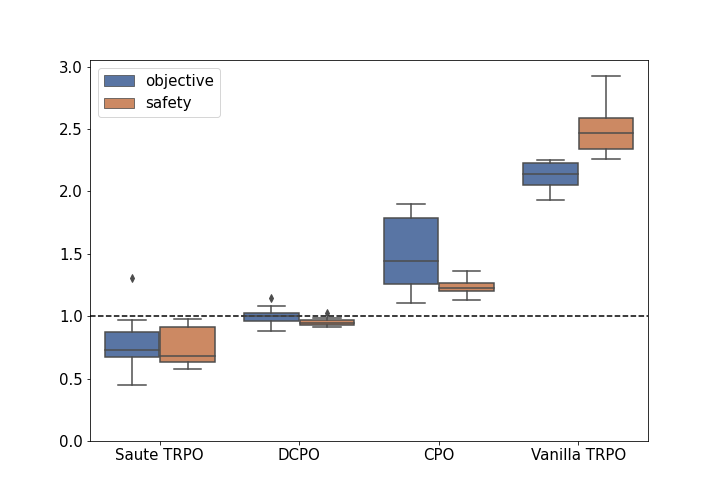}
  \caption{Normalised returns and costs after 30 episodes for TRPO, Saute TRPO, DCPO, and CPO over 10 runs. The returns are divided by $329.5$ (the average performance of DCPO) and costs by $15.0$ (the safety budget; marked by dashed line).}
  \label{fig:final_eval}
\end{figure}

Figure \ref{fig:final_eval} shows box plots for the normalised costs (safety) and returns (objective) seen in testing the trained algorithms for $30$ epochs. The results seen during inference are analogous to those at the end of training and are detailed in Figure \ref{tab:final_evaluation}. Indeed, DCPO shows the smallest variance in both the returns and the costs, with a decrease in the standard deviation with respect to CPO of $52.1\%$ for the costs and $74.8\%$ for the returns. Moreover, DCPO presents average reliable constraint satisfaction within a standard deviation from $d$, albeit with some outlying constraint-violating trajectories. Nevertheless, it does a better job at satisfying constraints than CPO, which has absolutely no trajectories that do so. This suggests that the case study here is perhaps more challenging (in terms of constraints) than the environments on which CPO is better-suited. 
Furthermore, this suggests that DCPO is particularly beneficial for environments where constraint satisfaction is a dominant consideration.

\begin{figure}[h]
    \begin{minipage}{0.5\textwidth}
        \centering
        \medskip
        \begin{tabular}{c | c  c}
        & $J_R\left(\pi^*\right)$ & $J_C\left(\pi^*\right)$ \\
        \hline \hline
        Vanilla TRPO & $\mathbf{2.123}$ & $2.493$ \\
        Saute TRPO & $0.786$ & $\mathbf{0.757}$ \\
        CPO & $1.499$ & $1.232$\\
        DCPO & $1.000$ & $0.953$
        \end{tabular}
        \captionof{table}{Mean.}
        \vspace{0.1in}
        \begin{tabular}{c | c  c}
        & $J_R\left(\pi^*\right)$ & $J_C\left(\pi^*\right)$ \\
        \hline \hline
        Vanilla TRPO & $0.112$ & $0.198$ \\
        Saute TRPO & $0.231$ & $0.162$ \\
        CPO & $0.298$ & $0.071$\\
        DCPO & $\mathbf{0.075}$ & $\mathbf{0.034}$
        \end{tabular}
        \captionof{table}{Standard deviation.}
    \end{minipage}
    \caption{Normalised evaluation performance for the discounted returns $J_R\left(\pi^*\right)$ and costs $J_C\left(\pi^*\right)$. Returns are normalised by dividing by $329.5$ (average performance of DCPO) and the costs by the safety budget $15.0$.}
    \label{tab:final_evaluation}
\end{figure}

\section{Conclusions}
This paper introduces an approach for trust region optimization in CPO, by reshaping episodic discounted costs and the maximum allowed value using the agent's confidence in constraint satisfaction, in turn given by an approximated cost distribution. 
The developed algorithm, which we call DCPO, manages to balance the trade-off between maximizing returns and minimizing the costs in a computational supply chain case study, such that constraints are satisfied (within one standard deviation) over ten replications. 
Nevertheless, DCPO allows for some constraint-violating trajectories, albeit to a much lower extent than CPO. Moreover, the incurred costs and returns are also more predictable compared to other methods, with a surprising $52.1\%$ decrease in variance in cost at the end of training compared to CPO, and a $74.8\%$ decrease for rewards (see Figure \ref{tab:final_evaluation}). 
Future work can explore ways to use the learned cost/return distributions to directly enforce constraints during training, rather than modifying the Lagrange multiplier of the safety constraints. It would also be interesting to explore different ways to parameterise the cost/value distributions, possibly by using the QR-DQN \cite{quantile-regression} approach of learning the positions of the quantiles rather than having these fixed. 

\vspace{-0.2cm}
\section{Acknowledgments}
The authors acknowledge support from the Engineering \& Physical Sciences Research Council (EPSRC) through fellowship EP/T001577/1 and an Imperial College Research Fellowship to CT.

% ================================ references ===================================
%% References 

\bibliographystyle{IEEEtran}
\bibliography{references}

\begin{thebibliography}{10}
\providecommand{\url}[1]{#1}
\csname url@samestyle\endcsname
\providecommand{\newblock}{\relax}
\providecommand{\bibinfo}[2]{#2}
\providecommand{\BIBentrySTDinterwordspacing}{\spaceskip=0pt\relax}
\providecommand{\BIBentryALTinterwordstretchfactor}{4}
\providecommand{\BIBentryALTinterwordspacing}{\spaceskip=\fontdimen2\font plus
\BIBentryALTinterwordstretchfactor\fontdimen3\font minus
  \fontdimen4\font\relax}
\providecommand{\BIBforeignlanguage}[2]{{%
\expandafter\ifx\csname l@#1\endcsname\relax
\typeout{** WARNING: IEEEtran.bst: No hyphenation pattern has been}%
\typeout{** loaded for the language `#1'. Using the pattern for}%
\typeout{** the default language instead.}%
\else
\language=\csname l@#1\endcsname
\fi
#2}}
\providecommand{\BIBdecl}{\relax}
\BIBdecl

\bibitem{nian2020review}
R.~Nian, J.~Liu, and B.~Huang, ``A review on reinforcement learning:
  Introduction and applications in industrial process control,''
  \emph{Computers \& Chemical Engineering}, vol. 139, p. 106886, 2020.

\bibitem{shin2019reinforcement}
J.~Shin, T.~A. Badgwell, K.-H. Liu, and J.~H. Lee, ``Reinforcement
  learning--overview of recent progress and implications for process control,''
  \emph{Computers \& Chemical Engineering}, vol. 127, pp. 282--294, 2019.

\bibitem{ikonen2020reinforcement}
T.~J. Ikonen, K.~Heljanko, and I.~Harjunkoski, ``Reinforcement learning of
  adaptive online rescheduling timing and computing time allocation,''
  \emph{Computers \& Chemical Engineering}, vol. 141, p. 106994, 2020.

\bibitem{shyalika2020reinforcement}
C.~Shyalika, T.~Silva, and A.~Karunananda, ``Reinforcement learning in dynamic
  task scheduling: A review,'' \emph{SN Computer Science}, vol.~1, pp. 1--17,
  2020.

\bibitem{garcia2015comprehensive}
J.~Garc{\i}a and F.~Fern{\'a}ndez, ``A comprehensive survey on safe
  reinforcement learning,'' \emph{Journal of Machine Learning Research},
  vol.~16, no.~1, pp. 1437--1480, 2015.

\bibitem{distributional-perspective}
M.~G. Bellemare, W.~Dabney, and R.~Munos, ``A distributional perspective on
  reinforcement learning,'' in \emph{International Conference on Machine
  Learning}.\hskip 1em plus 0.5em minus 0.4em\relax PMLR, 2017, pp. 449--458.

\bibitem{petsagkourakis2022chance}
P.~Petsagkourakis, I.~O. Sandoval, E.~Bradford, F.~Galvanin, D.~Zhang, and
  E.~A. del Rio-Chanona, ``Chance constrained policy optimization for process
  control and optimization,'' \emph{Journal of Process Control}, vol. 111, pp.
  35--45, 2022.

\bibitem{bellman-dp}
R.~Bellman, \emph{{Dynamic Programming}}.\hskip 1em plus 0.5em minus
  0.4em\relax Dover Publications, 1957.

\bibitem{policy-gradients}
R.~S. Sutton, D.~McAllester, S.~Singh, and Y.~Mansour, ``Policy gradient
  methods for reinforcement learning with function approximation,'' in
  \emph{Proceedings of the 12th International Conference on Neural Information
  Processing Systems}, ser. NIPS'99.\hskip 1em plus 0.5em minus 0.4em\relax
  Cambridge, MA, USA: MIT Press, 1999, p. 1057–1063.

\bibitem{GAE}
J.~Schulman, P.~Moritz, S.~Levine, M.~Jordan, and P.~Abbeel, ``High-dimensional
  continuous control using generalized advantage estimation,''
  \emph{arXiv:1506.02438}, 2015.

\bibitem{trpo}
J.~Schulman, S.~Levine, P.~Abbeel, M.~Jordan, and P.~Moritz, ``Trust region
  policy optimization,'' in \emph{International Conference on Machine
  Learning}.\hskip 1em plus 0.5em minus 0.4em\relax PMLR, 2015, pp. 1889--1897.

\bibitem{yoo2022dynamic}
H.~Yoo, V.~M. Zavala, and J.~H. Lee, ``A dynamic penalty approach to state
  constraint handling in deep reinforcement learning,'' \emph{Journal of
  Process Control}, vol. 115, pp. 157--166, 2022.

\bibitem{cpo}
J.~Achiam, D.~Held, A.~Tamar, and P.~Abbeel, ``Constrained policy
  optimization,'' in \emph{International Conference on Machine Learning}.\hskip
  1em plus 0.5em minus 0.4em\relax PMLR, 2017, pp. 22--31.

\bibitem{saute}
A.~Sootla, A.~I. Cowen-Rivers, T.~Jafferjee, Z.~Wang, D.~H. Mguni, J.~Wang, and
  H.~Ammar, ``Saut{\'e} {RL}: Almost surely safe reinforcement learning using
  state augmentation,'' in \emph{International Conference on Machine
  Learning}.\hskip 1em plus 0.5em minus 0.4em\relax PMLR, 2022, pp.
  20\,423--20\,443.

\bibitem{inventory-management}
H.~D. Perez, C.~D. Hubbs, C.~Li, and I.~E. Grossmann, ``Algorithmic approaches
  to inventory management optimization,'' \emph{Processes}, vol.~9, no.~1,
  2021.

\bibitem{HubbsOR-Gym}
C.~D. Hubbs, H.~D. Perez, O.~Sarwar, N.~V. Sahinidis, I.~E. Grossmann, and
  J.~M. Wassick, ``Or-gym: A reinforcement learning library for operations
  research problems,'' \emph{arXiv:2008.06319}, 2020.

\bibitem{risk-uncertainty}
W.~R. Clements, B.~Van~Delft, B.-M. Robaglia, R.~B. Slaoui, and S.~Toth,
  ``Estimating risk and uncertainty in deep reinforcement learning,''
  \emph{arXiv:1905.09638}, 2019.

\bibitem{quantile-regression}
W.~Dabney, M.~Rowland, M.~G. Bellemare, and R.~Munos, ``Distributional
  reinforcement learning with quantile regression,'' \emph{arXiv:1710.10044},
  2017.

\end{thebibliography}

\end{document}